# Deep Learning-based Hybrid Graph-Coloring Algorithm for Register Allocation

Dibyendu Das
Consultant
dibyendu.das0708@gmail.com

Shahid Asghar Ahmad
AMD
Asghar-ahmad.Shahid@amd.com

Kumar Venkataramanan
AMD
Venkataramanan.Kumar@amd.com

**ABSTRACT**

Graph-coloring is an NP-hard problem which has a myriad of applications. Register allocation, which is a crucial phase of a good optimizing compiler, relies on graph coloring. Hence, an efficient graph-coloring algorithm is of paramount importance. In this work we try to 'learn' a good heuristic for coloring interference graphs that are used in the register allocation phase. We aim to handle moderate-sized interference graphs which have 100 nodes or less. For such graphs we can get the optimal allocation of colors to the nodes. Such optimal coloring is then used to train our Deep Learning (DL) network which is based on several layers of LSTM that output a color for each node of the graph. However, the current network may allocate the same color to the nodes connected by an edge resulting in an invalid coloring of the interference graph. Since it is difficult to encode constraints in an LSTM to avoid invalid coloring, we augment our deep learning network with a **color correction** phase that runs after the colors have been allocated by the DL network. Thus, our algorithm is hybrid in nature consisting of a mix of a DL algorithm followed by a more traditional correction phase. The color correction phase handles the edges with invalid coloring by first trying to reuse a color allocated to other nodes that are not connected to the invalid nodes, failing which it adds a totally new color – thereby breaking the invalid allocation. Our experience with many graphs shows that around 10%-30% edges may get an invalid coloring. We have trained our DL network using several thousand random graphs of varying sparsity(density). On application of our hybrid algorithm to various popular graphs found in literature we see that our algorithm does very well when compared to the optimal coloring of these graphs. We have also run our algorithm against LLVM's popular greedy register allocator for several SPEC CPU® 2017 benchmarks and notice that the hybrid algorithm performs on par or better than such a well-tuned allocator for most of these benchmarks.

## 1 Introduction

For solving NP-hard problems like graph-coloring, numerous heuristics have been designed. Comprehensive reviews of such heuristics can be found in [13,24]. Two well-known greedy algorithms DSATUR [5,7] and RLF [21] employ refined rules to dynamically determine the next vertex to color. These greedy heuristic algorithms are usually fast. Register allocation as a graph coloring problem was first introduced by Chaitin in [8]. Later, several other graph coloring register allocation algorithms have been introduced by Chow [9] and Briggs [3]. These heuristics achieve good performance over a wide range of interference graphs. However, there may be scope for improvement in terms of optimizing the number of registers used and reducing the cost of spilling registers to memory. While one can pursue designing smarter heuristics our goal is to learn a good heuristic using DL techniques that will be as close to the optimal assignment as possible and compare favorably, if not surpass, some of the heuristics used in modern register allocators like LLVM.

In order to learn a good heuristic for graph coloring with interference graphs in mind, we need training data that includes solutions that outperform existing heuristics in the register allocators. We tackle this problem firstly by restricting the interference graph size to a maximum of 100 nodes. This is a reasonable number based on our experience of working with the LLVM register allocator. Secondly, we use an exact algorithm to solve the graph coloring problem for such graphs having less than 100 nodes. It has been found that such exact solvers work well for graphs having small to moderate size. In this work we have found that such a method takes a long time to color a graph having more than 75 nodes. However, since this coloring is done for DL training the additional time becomes acceptable.

In this paper, we demonstrate our approach by introducing a hybrid algorithm that consists of a deep learning-based technique augmented with a color correction phase. We use a Recurrent Neural Network (RNN) [14], specifically an LSTM (Long short-term memory) [18], to model the graph coloring problem. Constraints like the nodes appearing at the ends of an edge should not have the same color, cannot be encoded well in an LSTM. As a result, despite a large training set, the LSTM may not learn such a constraint fully. To compensate for this, we have designed a traditional post-pass after LSTM, that corrects this anomaly by checking all such invalid edges.

We train our LSTM using random graphs generated using the very_nauty [4] package. For inference we use popular graphs found in literature as well as interference graphs that can be generated by LLVM-9.0 [23] as part of its register allocation phase. Note that the random graphs generated and the interference graphs of LLVM may have different characteristics in terms of sparsity, node degrees, and other graph parameters. However, our observation is that our LSTM-based approach generalizes well from random graphs to interference graphs.

The main contributions of our paper are as follows:
- We have designed a new LSTM-based DL algorithm that can color graphs and the number of colors used compares favorably with optimal coloring
- The DL algorithm is paired with a color correction phase that corrects nodes which may have been colored in an invalid manner
- We show, with popular graphs and interference graphs culled from the LLVM-generated SPEC CPU® 2017 [29] benchmarks, that the efficacy of our approach is high

The paper is organized as follows. In Section 2 we discuss how we model the graph-coloring problem using LSTM. We also discuss how to train such models and provide the algorithm for color correction. In Section 3 we look at some popular graphs as use-cases and compare optimal coloring of such graphs with our method. Section 4 will deal with the interference graphs generated using LLVM and the performance of our algorithm on such graphs. We propose a possible architecture of how to incorporate a DL-based module in LLVM also. In Section 5 we will discuss related work. We conclude in Section 6 and discuss possible future Work.

## 2 Graph Coloring using LSTM

In this work we model the graph coloring problem using LSTM, which is a variant of Recurrent Neural Network (RNN). A common LSTM unit is composed of a cell, an input gate ($i$), an output gate ($o$) and a forget gate ($f$) as shown in Figure 1 [10]. The cell remembers values over arbitrary time intervals and the three gates regulate the flow of information into and out of the cell where $h$ is the hidden state and $X$ is the input, at time step $t$. The final value of $h_n$ is the output of the LSTM. LSTM networks are well-suited for classifying, processing and making predictions based on a data sequence which appears as a sequence of time steps and were developed to deal with the exploding and vanishing gradient problems that can be encountered when training traditional RNNs. A useful way to visualize RNNs/LSTMs is to consider the update graph formed by 'unfolding' the network along the input sequence. The unfolded/unrolled LSTM with multiple cells is shown in Figure 2 [1].

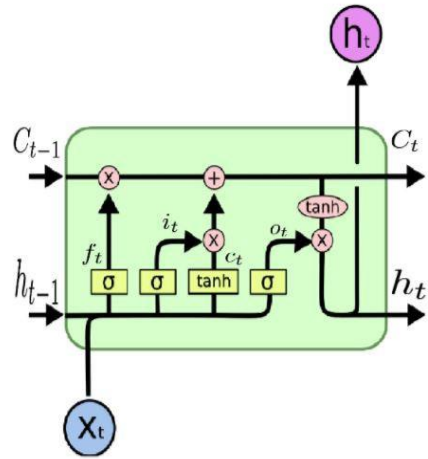

Figure 1: An LSTM cell

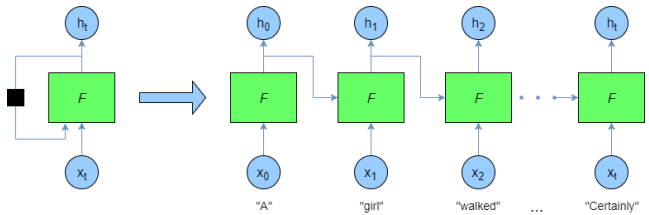

Figure 2: Unrolled/Unfolded LSTM/RNN

### 2.1 LSTM-based Model for Graph Coloring

For graph coloring, since we handle modest-sized graphs of 100 nodes or less, we use the entire adjacency matrix of a graph as input. The input sequence to the LSTM is the sequence of graph nodes – starting with node *0* and ending with node *n-1* where *n* is the number of nodes of the graph. For each node, we use the adjacency vector of the node as shown in Figure 3, as input. Hence, at each time step *t* of the input sequence to the LSTM, the adjacency vector of $v_t$ is provided, where $v_t$ is the *t*-th node of the graph. The adjacency vector of $v_t$ is nothing else but the entire row of the adjacency matrix corresponding to node $v_t$. The output of the LSTM-based model are the colors of each node. Since there are *n* vertices the output sequence is also of size *n*. For better performance and prediction accuracy we use deep LSTMs with 3 layers, with hidden states of one layer passed on to the LSTM cells of the next layer as inputs.

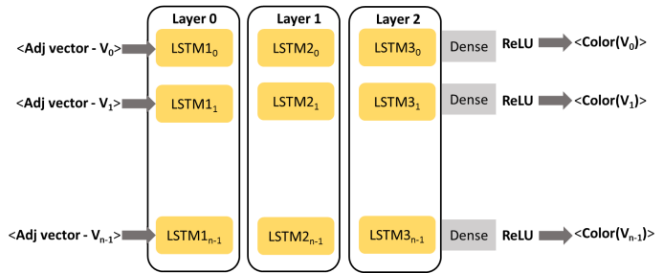

**Figure 3: Deep Learning Model for Graph Coloring using 3-layered LSTMs**

After the LSTM layers, we use a fully connected dense layer that takes the output of the final LSTM layer and produces a single value. This value is fed through a ReLU [14] to provide the final color value of a node. Since a graph of *n* nodes never requires more than *n* colors to correctly color all its vertices, we can use the numbers *1 ... n* to represent the colors of the vertices at the output. The three layers are numbered Layer 0, Layer 1 and Layer 2 in Figure 3. In each layer we show the unrolled LSTM consisting of the full sequence size *n* with $LSTM_0$ corresponding to input time step of *0* and $LSTM_{n-1}$ corresponding to the input time step of *n-1*, which corresponds to the vertices $v_0$ and $v_{n-1}$ of the graph. In our model, each LSTM cell has 1024 hidden units. The number of layers (3) as well as the number of hidden units in each cell has been arrived at empirically. We have coded the model using Tensorflow 2.0 [15].

**2.2 Training the Model**

For training our model we use random graphs generated via the very_nauty package [4]. Very_nauty is a C library of graph algorithms, especially targeted at very fast generation of random graphs, and exact clique number and chromatic number computations. In practice, it is possible to use the exact algorithms on graphs with up to a few hundred nodes. However, we noticed that with this software, exact graph coloring over 75 nodes becomes quite slow in practice though we can get the exact color allocation in a reasonable time up to 150 nodes or so. However, in this work we will restrict our input graphs to 100 nodes or less as mentioned earlier. We mainly use two functions from this package.

First, a function called *graph_gnp(graph_t g, double p)* which uses the Erdös-Renyi model [19] to generate random graphs for *n* nodes with *p* being the probability of two nodes being connected by an edge. Second, we use *graph_chromatic_number(graph_t g,clock_t timeout)* to compute the exact chromatic number of the graph from which we can also extract the coloring allocation to the individual nodes of the random graph that has been generated. For training, we have generated close to 10000 random graphs consisting of one node to one hundred nodes. For each such graph we vary the parameter *p* of *graph_gnp* between 0.05 to 0.95 implying very sparse to very dense graphs. For each such graph generated we use *graph_chromatic_number* function to find the optimal allocation of colors. It should be noted that we use only one optimal color allocation to guide the training. Once the colors are assigned optimally, we can permute those colors among the nodes to get other optimal color assignments. But we do not consider such instances as additional training samples. This is done to keep the training time manageable as we train on traditional CPU-based systems for this work.

**2.2.1 Input and Output formats for Training**

For training, the input to the model is a set of 10000 samples provided in a .csv file. In order to encode the full adjacency matrix, we encode the adjacency vector of each node sequentially. This sequence consists of ones or zeros up to maximum value of 100 time steps. In order to encode the 100-element adjacency vector in a compact manner we use 2 LONG INT values. Since each LONG INT can encode up to 64 bits of zeros and ones, we use 2 LONG INTS to encode 100 nodes of the adjacency vector. We use zero padding as necessary. Following the adjacency vector sequence of 2*100 nodes, we list out the colors of each node using values *1* to *n,* once again zero-padding as necessary. Thus, each sample of the training set consists of the following data: <Number of optimal colors used, **Input:** 200 LONG INTs (2 per vertex), **Output:** 100 colors (1 per vertex) >. The optimal coloring as provided in the output of the training sample is checked against the colors assigned by the model during each epoch and the error metric *mean_absolute_percentage_error* is used to compare the two. The compressed adjacency vectors are expanded to 128-element arrays of zeros or ones before they are fed as inputs to the first LSTM layer. We train the model for 100 epochs which achieves a training error of about 5%.

**2.2.2 Inference and Color Correction**

During the inference phase, the trained model is used to predict the colors assigned to each vertex of a new sample.

The input for inferencing is slightly different from the input for training as we do not require the output sequence of vertex colors to be provided. However, to keep matters simple we still retain the same input format as training but fill out the entire output sequence with ones. The input sequence is similar to that of training with 2 LONG INTs being used to code each adjacency vector.

Once the trained model predicts the colors of each node of a new graph, we test for the validity of the prediction. This implies checking each edge of the graph to see whether the two end points of the edge have the same color. If no such edge is found, the prediction is considered valid. Otherwise, the prediction is deemed invalid and we need to apply **color correction** to reach at a new color assignment for each such invalid edge, so that the end points have different colors.

```
Algorithm: ColorCorrection()
Colors = Set of colors allocated to the graph after inference
for each INVALID edge e=<n1,n2> do {
    Let c=color of the nodes n1 and n2
    for ( c1 in Colors AND c1 != c ) do {
        if ( exists e1 = <n1,M> such that M has color c1 )
            continue;
        Color c1 is not used by any neighbor of n1
        reuse color c1 for n1
    }
    if no color found for reuse for n1 then {
        for ( c2 in Colors AND c2 != c ) do {
            if exists e1 = <n2,M> such that M has color c2
                continue;
            Color c2 is not used by any neighbor of n1
            reuse color c2 for n1
        }
    }
    if no color found for reuse for both n1 AND n2 then {
        Create new color cn
        Colors = Colors U {cn}
        Assign cn to n1
    }
}
```

**Algorithm 1: Color Correction**

Color correction is a post-pass that runs after the inferencing phase is completed. Algorithm 1 listed here shows how this phase works. We examine each invalid edge $e=<n1,n2>$. For each such edge we first examine $n1$ and see whether the entire color palette allocated by the inference phase is exhausted by the neighbors of $n1$. If not, we can choose the first missing color and use it to color $n1$.

If this mechanism fails for $n1$ we test for $n2$. If both fail, we need to create a new color which is then added to the color palette and $n1$ is colored using the new color.

For inference. we generate a separate test set of about 7600 graphs using the random graph generator but using $p$ values which are slightly perturbed from those used in the training set. This is to test the robustness of the trained model.

For the test set we use the percentage of invalid edges as a metric and observe ~12% invalid edges over the entire test set on an average. Also, for ~17% of the cases the LSTM model predicts and allocates as well as the optimal coloring scheme. Which means that color correction is not required for these cases. For ~70% of cases our hybrid allocator uses not more than 3 extra colors when compared to the optimal coloring. We also observe that for ~9% cases our coloring requires 6 to 10 extra colors implying that for these cases our algorithm may be performing below par.

We demonstrate the working of the color correction algorithm using a popular graph from literature called the Forest-Fire graph [27]. This graph has 10 nodes, 18 edges and a chromatic number of 5. Our LSTM-based model colors aggressively with 4 colors resulting in 2 invalid edges as shown in Figure 4. These edges are $<v_2,v_3>$ where both the vertices carry the color $c_3$ and $<v_1,v_5>$ where both the vertices carry the color $c_2$. $v_5$ can reuse the color $c_3$ as none of its neighboring nodes use $c_3$. $v_2$ requires a new color $c_5$ as both $v_2$ and $v_3$'s neighbors use all the colors.

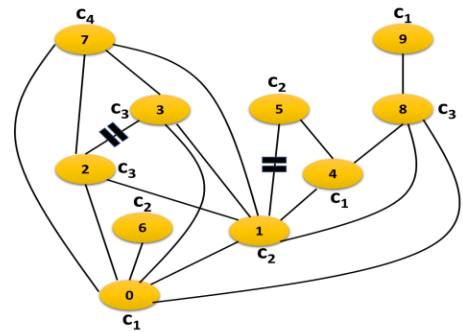

**Figure 4: Forest-Fire Graph with 2 invalid edges (crossed) after inference**

## 3 Performance on some popular graphs

In this section we will look at the performance of our model when applied to some popular graphs found in the literature. We saw in the Forest-Fire graph that after inferencing and color correction we were able to color the graph optimally with 5 colors which is also the chromatic number of the graph.

First, we will look at the Karate graph shown in Figure 5 [25]. This graph has 34 nodes and 78 edges, and its chromatic number is 5.

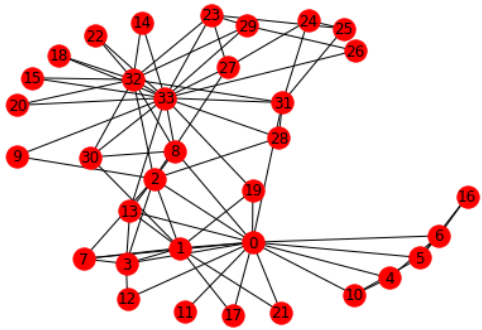

Figure 5: Karate Graph

For this graph after inferencing we find that the graph is colored using 4 colors and consists of 23 invalid edges out of 79 edges which is ~30% of the edges. On applying color correction, we use only one extra color resulting in coloring the graph optimally using 5 colors.

Second, we use the Chvatal graph [30] which consist of 12 nodes and 24 edges. The chromatic number of the graph is 4.

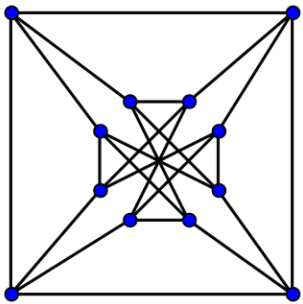

Figure 6: Chvatal Graph

For this graph after inferencing we find that the graph is colored using 3 colors and consists of 7 invalid edges out of 24 edges which is ~28% of the edges. On applying color correction, we use only one extra color resulting in coloring the graph optimally using 4 colors.

Third, we use the Baidu graph [20] which has been demonstrated for a new Reinforcement Learning-based approach to graph coloring by Baidu engineers. This graph consists of 60 nodes and 90 edges. The chromatic number of the graph is 3 but today's best graph-coloring heuristics can, at best, color the graph using 4 colors.

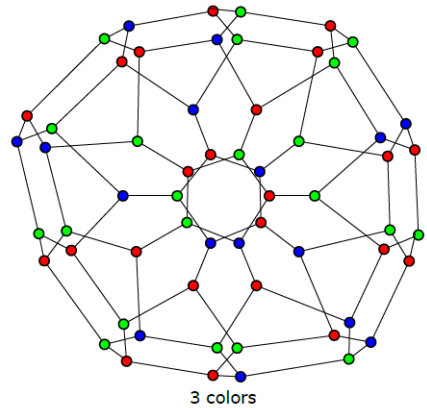

Figure 7: Baidu Graph

For this graph after inferencing we find that the graph is colored using 3 colors with 35 out of 90 edges being invalid which is ~38%. We are unable to reach the optimal number of 3 colors. But we can match the coloring number of the best heuristics available today by being able to color with 4 colors.

Fourth, we use a planar graph cited in the paper [16]. The graph has 31 nodes and 72 edges with a chromatic number of 3.

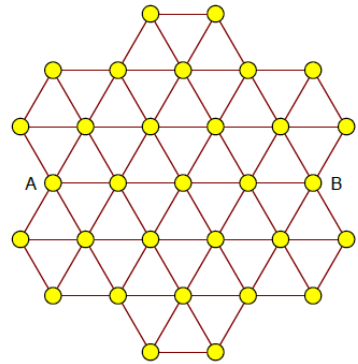

Figure 8: Planar Graph

Our inference model colors this graph with 4 colors after which 19 edges out of 72 are found to be invalid which is about 26%. Color correction adds an extra color resulting in the graph being colored using 5 colors. Thus, we consume 2 extra colors compared to the optimal coloring.

Finally, we look at a bunch of graphs culled from the COLOR02/03/04 workshop dataset [11] which lists many graphs with their structures and chromatic numbers. This data set is also used in [22]. We choose a few of these graphs having less than 100 nodes.

| Graph-name | <n,e> | X(G) | Predicted(G) before CC | Predicted(G) after CC | % invalid edges |
|---|---|---|---|---|---|
| insertions2 | 37,72 | 4 | 3 | 4 | 34% |
| insertions3 | 56,110 | 4 | 3 | 4 | 41% |
| insertions4 | 67,232 | 4 | 5 | 6 | 28% |
| mugg100 | 100,166 | 4 | 3 | 4 | 35% |
| queens8_12 | 96,1368 | 12 | 9 | 17 | 12% |
| queens5x5 | 25,160 | 5 | 6 | 9 | 21% |
| queens6x6 | 36,290 | 7 | 6 | 9 | 16% |
| queens7x7 | 49,476 | 7 | 7 | 11 | 15% |
| queens8x8 | 64,728 | 9 | 8 | 12 | 13% |

**Table I: COLOR dataset results**

Table I lists the results for some of the COLOR graphs having less than or equal to 100 nodes. χ(G) is the chromatic number, and the two following columns show the predicted colors before and after color correction (referred to as CC). The last column provides the number of invalid edges. From these results in Table I it appears that for sparser graphs, ex: insertions2,3 and mugg100 our model produces results which matches the optimal coloring number. For other graphs like queens8_12 which is denser our hybrid method requires 5 extra colors compared to the optimal coloring. Other results are somewhere in between.

## 4 Performance comparison with LLVM's Greedy Register Allocator (GRA)

Graph coloring register allocators construct an interference graph. Program values are represented by nodes (also called virtual registers) in the interference graph and edges between nodes imply that those values cannot share a physical register as their live ranges/intervals overlap. It is the allocator's responsibility to map the unlimited virtual registers into a finite number of machine registers.

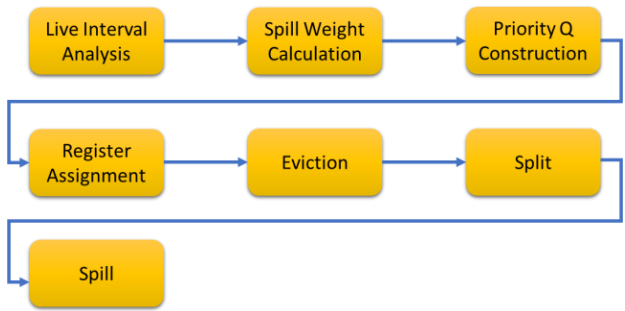

**Figure 9: LLVM Register Allocator Flow**

LLVM's register allocation (regalloc) pass is part of Codegen. The general flow of the pass is as shown in Figure 9 adapted from [32].

The default register allocator in LLVM is called the Greedy Register Allocator (GRA). GRA's approach is based on the live interval information of the program variables. Initially, the spill weight calculation of available live intervals is performed based on heuristics such as use density, rematerializability etc. A priority queue is constructed and populated with these live intervals based on the scope of the program variables. Globals are given higher and locals given lower priority. Higher priority intervals are picked from the priority queue and assigned to available physical registers. In case of non-availability of physical registers and/or interferences between live intervals various approaches such as eviction, splitting and spilling of live intervals are employed (collectively called selection heuristics) to find the allocation or coloring. Eviction is the process of changing an assigned interval to unassigned based on a lower spill cost. Splitting is the process of dividing a chosen live interval into smaller intervals in case of a failed eviction. Eviction and splitting are performed while keeping the priority queue updated with the victims of eviction and split live intervals. If eviction and splitting fail, spilling of intervals are employed. Split and spill may create new live ranges which are put back in the priority queue though for simplicity that interaction is not shown in Figure 9. More information about regalloc can be found in [6].

GRA does not maintain an interference graph explicitly. Hence, we create the interference graph at the end of the Live Interval Analysis phase. Initially, all the intervals are added to the interference graph as nodes and then the edges are added iteratively while checking if two live intervals overlap. The interference graph is then written out in the input format required for inferencing as outlined in Section 2.2.2 in a .csv file. We collect the interference graphs for the functions of certain SPEC CPU® 2017 [29] benchmarks. We ignore those functions which have more than 100 nodes.

In order to compare the register allocation quality of our hybrid model with the final allocation done by the complete register allocator of LLVM we also count the exact number of unique registers used by each function of a SPEC® benchmark after code generation. At the end of register allocation phase, LLVM provides a mapping between virtual and physical registers. For all the virtual registers, we scan this map and extract the physical registers and count them uniquely. For architectures like x86, registers AH, AX, EAX and RAX share the same physical location, but they have different sizes. LLVM represents these physical registers as register units or sub registers, where each unit is an alias. We also take care of this and do not count registers that alias with each other as separate colors/registers.

**4.1 Results from some SPEC CPU® 2017 benchmarks**

In this study, we collect the interference graphs of a large set of functions from the following SPEC CPU® 2017 benchmarks. The benchmarks are *505.mcf_r*, *557.xz_r*, *541.leela_r*, *508.namd_r* and *502.gcc_r*. These benchmarks have been compiled using the LLVM-9 compiler [23] at an optimization level of -O3 and the interference graphs collected. We use inferencing on these graphs and get the color allocation and predictions – both before and after the color correction phase. We also compare these values with the number of registers used by these functions as allocated by GRA.

*505.mcf_r*

In Table II, we list some of the mcf functions having interference graphs of 100 nodes or less and their coloring numbers.

| Functions | LLVM reg-alloc | DL before correction | DL after correction |
|---|---|---|---|
| switch_arcs | 17 | 14 | 22 |
| replace_weaker_arc | 16 | 10 | 13 |
| insert_new_arc | 14 | 11 | 15 |
| resize_prob | 7 | 4 | 7 |
| marc_arcs | 12 | 10 | 12 |
| refreshPositions | 14 | 14 | 14 |
| refreshArcPositions | 8 | 4 | 7 |
| master | 23 | 13 | 24 |
| worker | 25 | 16 | 24 |
| markBaskets | 11 | 9 | 9 |
| primal_bea_mpp | 21 | 14 | 26 |
| primal_feasible | 9 | 10 | 10 |
| flow_org_cost | 14 | 7 | 10 |
| flow_cost | 13 | 8 | 10 |
| refresh_neigbour_lists | 10 | 6 | 9 |
| update_tree | 19 | 8 | 19 |
| primal_start_artificial | 11 | 7 | 9 |
| primal_imnus | 7 | 5 | 7 |
| write_objective_value | 5 | 5 | 8 |
| main | 6 | 3 | 4 |
| TOTAL | 262 | 174 | 257 |

**Table II: 505.mcf_r results**

Our hybrid method outperforms GRA by a small margin of ~2%. However, please note that our allocator does not inspect the types of variables and register classes for allocation (ex: whether a vector data-type does uses a scalar register) and hence may be slightly more aggressive than LLVM's GRA. Our basic DL model is quite aggressive and allocates ~35% fewer registers than GRA though for certain functions ex: *refreshPositions* or *write_objective_value* it matches the number of registers used by GRA. For *update_tree* the DL model uses only 8 colors compared to GRA's 19 but after color correction the number comes back up to 19 suggesting that for some graphs that are generated by LLVM, our initial model does not do an adequate job.

*557.xz_r*

In the following table, Table IV, we list some of the functions of xz having interference graphs of 100 nodes or less and their coloring numbers.

| Functions | LLVM reg-alloc | DL before correction | DL after correction |
|---|---|---|---|
| lzma_index_buffer_decode | 11 | 7 | 9 |
| index_decode | 12 | 13 | 15 |
| lzma_index_hash_append | 10 | 10 | 10 |
| lzma_index_hash_decode | 14 | 16 | 18 |
| lzma_stream_buffer_decode | 14 | 6 | 11 |
| stream_decode | 14 | 19 | 19 |
| lzma_stream_footer_decode | 8 | 6 | 6 |
| lzma_vli_decode | 13 | 8 | 13 |
| lz_encoder_prepare | 13 | 6 | 13 |
| lzma_lz_encoder_init | 11 | 7 | 9 |
| lz_encode | 14 | 10 | 15 |
| lzma_mf_hc3_find | 21 | 9 | 19 |
| … | … | … | … |
| TOTAL | 732 | 500 | 715 |

**Table III: 557.xz_r results**

Since we have tracked many more functions that can be shown in the table, we list only a few but the last row shows the overall total of all the functions we have tracked. For xz our allocator is ~2.5% better than GRA, showing behavior like the benchmark mcf in terms of uplift.

*508.namd_r*

In the following table, Table III, we list some of the functions of namd. Namd has quite a few functions having more than 100 nodes – some of them running to over 500 nodes or so. Also, namd is the only benchmark (among the 5 we investigated) where GRA performs ~5% better than our allocator for the functions we investigated, mainly due to:*_ZN9ResultSet8readfileEP8_IO_FILE,_Z5equalPdS_S_ S,_ZN5Patch8readfileEP8_IO_FILEP8Molecule*.

On investigating the graph structures of these functions, we find that all of them have a few nodes which have very high degrees that is almost equal to the size of the graphs (40-60 nodes) while the rest have degrees of 4-6 implying that our model may not have trained well for such skewed graphs.

| Functions | LLVM reg-alloc | DL before correction | DL after correction |
|---|---|---|---|
| _ZN7LJTable8readfileEP8_IO_FILE | 15 | 9 | 15 |
| _ZN11ComputeList8readfileEP8_IO_FILEP9PatchList | 13 | 5 | 11 |
| _ZN26ComputeNonbondedWorkArraysC2Ev | 7 | 4 | 6 |
| _ZN26ComputeNonbondedWorkArraysD2Ev | 4 | 3 | 3 |
| _ZN11SelfCompute6doWorkEP9PatchList | 7 | 6 | 8 |
| _ZN8Molecule8readfileEP8_IO_FILE | 13 | 8 | 16 |
| _ZN11ObjectArenaIcE11getNewArrayEi | 8 | 3 | 6 |
| _ZN11ResizeArrayIPcE3addERKS0_ | 13 | 7 | 10 |
| _ZN5Patch9moveatomsEv | 11 | 7 | 12 |
| _ZN5Patch8readfileEP8_IO_FILEP8Molecule | 15 | 8 | 22 |
| _ZN9PatchList10setresultsEP9ResultSet | 21 | 10 | 20 |
| _ZN9PatchList8readfileEP8_IO_FILEP8Molecule | 11 | 6 | 10 |
| _ZN9ResultSet9writefileEP8_IO_FILE | 10 | 13 | 13 |
| _ZN9ResultSet8readfileEP8_IO_FILE | 7 | 13 | 13 |
| _Z5equalPdS_S_S_ | 7 | 13 | 13 |
| _ZN9ResultSet5checkEv | 3 | 3 | 4 |
| _ZN9ResultSet7compareERS_ | 13 | 9 | 11 |
| _ZN13SimParameters8readfileEP8_IO_FILE | 9 | 7 | 8 |
| _ZN11ResizeArrayI6VectorE6resizeEi | 12 | 9 | 9 |
| Total | 199 | 143 | 210 |

Table IV: 508.namd_r results

*541.leela_r*

Among all the benchmarks we studied, leela demonstrates the best performance for our allocator when compared to GRA. We tracked around 80 functions of leela, hence we show only a short list of functions in Table V. For leela, our hybrid algorithm improves on GRA by more than 7%. Also, there are a few functions like _ZN9UCTSearch13dump_analysisEv where our allocator uses much lesser registers (5 vs 11) compared to GRA. On inspecting the adjacency matrix of the interference graph of this function it appears that the graph has 44 nodes and 64 edges resulting in an average degree close to 3. It is likely that the sparsity of the graph helps our algorithm produce better results than GRA.

| Functions | LLVM reg-alloc | DL before correction | DL after correction |
|---|---|---|---|
| _ZN9FullBoard12update_boardEii total colors | 12 | 13 | 16 |
| _ZN9FullBoard15predict_ko_hashEii | 16 | 6 | 14 |
| _ZN7KoStateC2ERKS_ | 11 | 4 | 7 |
| _ZNSt6vectorIySaIyEE13_M_insert_auxEN9__gnu_cxx17__normal_iteratorIPyS1_EERKy | 12 | 5 | 9 |
| _ZN7Playout3runER9FastStateb | 12 | 10 | 14 |
| _ZN7Playout20do_playout_benchmarkER9GameState | 13 | 5 | 8 |
| _ZN7Playout8mc_ownerER9FastStatei | 14 | 9 | 19 |
| _ZN9UCTSearch15play_simulationER7KoStateP7UCTNode | 14 | 8 | 10 |
| _ZN9UCTSearch6get_pvB5cxx11ER9GameStateR7UCTNode | 12 | 6 | 10 |
| _ZN9UCTSearch13get_best_moveEi | 13 | 9 | 10 |
| _ZN9UCTSearch13dump_analysisEv | 11 | 2 | 5 |
| _ZN7KoStateC2ERKS_ | 11 | 4 | 7 |
| _ZN9UCTSearch5thinkEii | 12 | 7 | 12 |
| ... | ... | ... | ... |
| TOTAL | 961 | 559 | 894 |

Table V: 541.leela_r results

*502.gcc_r*

We took a sample of 50 functions of varying sizes from this benchmark which has thousands of functions. Without listing out a table, we observe that our hybrid algorithm performs about 5% better than GRA.

## 4.2 An architecture of a DL-based register allocator

In this section we will look at how to fit in a DL-based allocator in the LLVM GRA flow. Such allocators are still very early in development and may not outperform GRA for all interference graphs. Hence, we propose a dual strategy whereby an interference graph is fed to the normal flow, as well as a parallel module that is a DL-based inference engine like our hybrid algorithm. The overview of the new design is provided in Figure 10 which creates a parallel pipeline to the Register Assignment phase and calls our hybrid allocator. Later the better allocation of the two is chosen. If the number of colors exceed the number of available registers, then eviction, split and spill are applied as required for the DL-based allocator too. If the allocation of GRA is found to be better than the DL-based one, the inference graph as well as the exact allocation is stored in a training database that can be used to augment an offline training of the LSTM-based model. Note that, GRA's Register Assignment and the following phases are iterative in nature whereby the top priority live-range is chosen, register assignment tried and the follow-up steps carried out. But our DL-based allocator produces all the colors in one go. To handle this difference, we propose to use the priority queue of live ranges, pick in order and assign the same register to all the live ranges to which the DL-based algorithm has assigned the same color. Which means, though we pick one live range to assign, we may assign the same register to multiple live ranges. These live ranges are removed from the queue. If we have additional live ranges remaining after all the registers are utilized, we pass on the remaining live ranges to the eviction, split and spill phase and would need to invoke the tradition register assignment. These interactions are shown in Figure 10.

Since we base our training on random graphs it is very likely that the training samples may not encompass all kinds of interference graphs that can be generated by compilers. Hence, storing details of graphs for which our allocator does worse compared to GRA, acts as a continuous learning mechanism. It should be noted that the DL-based engine is a python-based module that needs to consume the input interference graph in a .csv format. In addition, the output of this engine should be consumed by the later phases. At present this design is just a prototype and has not been implemented in the Codegen phase of LLVM.

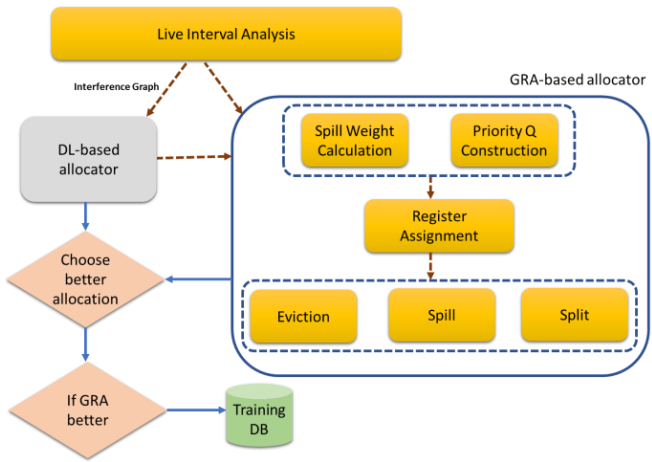

**Figure 10: LLVM Register Allocator Flow with DL allocator**

### 4.3 Some observations on our model

While training our LSTM-based model we mentioned the use of random graphs created by the very_nauty package – the graphs being based on the Erdös-Renyi (ER) [19] model. During inferencing we observe that for graphs generated by LLVM the LSTM-based model (before color correction) predicts 30%-40% lesser colors on an average when compared to GRA. And in some individual cases the difference is higher. We do not observe such behavior for the popular graphs mentioned in Section 3. One of the reasons is that the GRA is not an optimal algorithm and our model is trained on optimal allocation. However, this alone probably does not explain the full gap. On a closer look at some of the interference graphs generated by LLVM where we observe significant differences between the GRA and our allocator, we found that many of these graphs have skewed structures rather than regular ones. This means that few nodes of the graph have high degrees and connectivity while the rest have low or moderate degrees and connectivity. This implies that these graphs may lie in the class of scale-free networks [31] where the degree distribution of the nodes follow the power law rather than a uniform distribution which is a characteristic of the ER model. The interference graphs from LLVM appear to be a mixed bag of regular and scale-free graphs and hence we may need to devise an ensemble model [14] where we train on regular graphs as well as scale-free graphs but using different LSTMs as shown in Figure 11. During inference, we feed the data through both the models and pick the one which provides more optimal allocation.

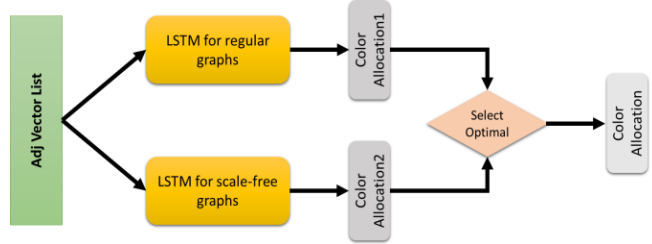

**Figure 11: Ensemble Model**

## 5 Related Work

There have been two recent works which deal with graph coloring using deep learning. The first is one from Baidu [20]. In this work the goal is to use deep reinforcement learning to color large graphs, as optimal coloring on such graphs is not possible with modern machines. Their work is inspired by AlphaGoZero [28] on HPC systems and use it to learn new graph coloring heuristics that improve the state-of-the-art accuracy by up to 10%. They can color graphs up to thousands of nodes. However, the training hardware required to build such a coloring network is extremely complex and requires hundreds of GPUs. In addition, the algorithm itself has many steps and not amenable to easy understanding. In contrast, we concentrate on graphs created during register allocation which generally does not exceed a few hundred nodes. Our algorithm is much simpler to understand and implement and can be trained on CPUs without requiring a complex setup. The second work we refer to is by Lemos *et al.* [22] that uses Graph Neural Networks (GNNs) [26,33]. Usually, graph neural network models assign multidimensional representations, or embeddings, to vertices and edges. These embeddings are then refined according to some adjacency information throughout a given number of message-passing iterations. The adjacency information controls which are the valid incoming messages for a given vertex (or edge), these filtered messages undergo an aggregating function and finally a Recurrent Neural Network (RNN) receives the aggregated messages and computes the embedding update for the given vertex. Lemos *et al.* compares their model against several popular approaches like Tabucol [17] and greedy heuristics. However, their algorithm does not find a correct assignment of colors as they frequently color the graphs with colors lower than the chromatic number of a graph. Hence their algorithm cannot be used in scenarios like register allocation.

# 6 Conclusion and Future Work

In this paper we have shown how to apply a deep learning framework to color graphs with special emphasis on solving the register allocation problem. Our algorithm is hybrid in nature, as it consists of a post-pass color correction phase that follows an LSTM-based deep learning model. We show the performance of this algorithm on several popular graphs and on interference graphs generated by LLVM for several SPEC CPU® 2017 benchmarks and demonstrate that our algorithm compares favorably with either an optimal allocation or state-of-the-art heuristics that have been tuned for quite some time.

To utilize deep learning-based graph coloring register allocation in production compilers we still need to carry out further experiments and studies. One of them is to investigate on how the interference graphs differ from random graphs and how to incorporate such graphs into the training cycle. It is also important to check interference graphs from other compilers like gcc (GNU C/C++ compiler) or icc (Intel C/C++ compiler). Though training via random graphs fares favorably when used for interference graphs, we may still need to build ensemble models as described in Sec 4.3 to make the models more effective.

Our model takes adjacency vectors of the nodes as inputs. The sequence in which these nodes are fed to the input LSTM is solely dependent on the numbering of the nodes. This can probably be enhanced by first carrying out a breadth-first-search (BFS) on the graph and feeding the BFS sequence to the LSTM – instead of one based on node numbering. In addition, we can also experiment with bi-directional LSTMs to capture relationships between nodes and edges which may not be captured by uni-directional LSTM alone.

In the current work we have handled interference graphs of size 100 nodes or less. We will need to extend our LSTM-based model for larger graphs though based on our study of the SPEC benchmarks we did not see graphs bigger than few hundreds of nodes. In general LSTMs may not work well for very long sequences though there have been positive results using pre-training for sequences up to several thousand nodes [12]. Also, modern attention-based LSTMs can probably handle longer sequences better [2]. In addition to the issue of handling long sequences for LSTMs for bigger interference graphs, we need to find optimal coloring algorithms for such graphs for supervised training. As mentioned earlier, optimal coloring for large graphs is infeasible today. Hence, we may need to fall back on good heuristics for such graphs and train on these. An alternative to additional training or handling long LSTM sequences, is to partition a larger graph into graphs of size 100 nodes or less. We can color these subgraphs using our hybrid mechanism. Then, color correction can be applied to the inter-subgraph edges as applicable.

To conclude, this work is one of the first steps to replace hand-designed heuristics for register allocation via graph-coloring using a machine learning model. As we learn and understand more, both about the applicable models and about the nature of interference graphs, we think that we may need to depend less and less on the color correction step and create a more powerful deep learning-based algorithm that can be used in future production compilers.